\def\year{2015}
\documentclass[letterpaper]{article}
\usepackage{aaai}
\usepackage{times}
\usepackage{bm}
\usepackage{algorithm}
\usepackage{algorithmic}
\usepackage{array}
\usepackage{amssymb}
\usepackage[cmex10]{amsmath}
\usepackage{amsthm}
\usepackage{amsfonts}
\usepackage{helvet}
\usepackage{graphicx}
\usepackage{caption}
\usepackage{courier}

\newcommand{\qihao}{\fontsize{8pt}{\baselineskip}\selectfont}
\newcommand{\bb}[1]{\raisebox{-3ex}[0pt][0pt]{\shortstack{#1}}}

\begin{document}
%
\title{Efficient Face Alignment via Locality-constrained Representation for Robust Recognition}
\author{Yandong Wen\textsuperscript{\textnormal{1}*}, Weiyang Liu\textsuperscript{\textnormal{2}*}, Meng Yang\textsuperscript{\textnormal{3}}, and Zhifeng Li\textsuperscript{\textnormal{4}}\\
\textsuperscript{1}School of Electronic and Information Engineering, South China University of Technology\\
\textsuperscript{2}School of Electronic and Computer Engineering, Peking University\\
\textsuperscript{3}School of Computer Science and Software Engineering, Shenzhen University\\
\textsuperscript{4}SIAT, Chinese Academy of Sciences\ \ \ \ \\
}
\maketitle
\begin{abstract}
\begin{quote}
Practical face recognition has been studied in the past decades, but still remains an open challenge. Current prevailing approaches have already achieved substantial breakthroughs in recognition accuracy. However, their performance usually drops dramatically if face samples are severely misaligned. To address this problem, we propose a highly efficient misalignment-robust locality-constrained representation (MRLR) algorithm for practical real-time face recognition. Specifically, the locality constraint that activates the most correlated atoms and suppresses the uncorrelated ones, is applied to construct the dictionary for face alignment. Then we simultaneously align the warped face and update the locality-constrained dictionary, eventually obtaining the final alignment. Moreover, we make use of the block structure to accelerate the derived analytical solution. Experimental results on public data sets show that MRLR significantly outperforms several state-of-the-art approaches in terms of efficiency and scalability with even better performance.
\end{quote}
\end{abstract}
\section{Introduction}
Over the past years, face recognition has been and is still one of the most important and fundamental computer vision problem. Significant progresses have been made in face recognition, ranging from the family of sparse representation \cite{wright2009robust,wagner2012toward,zhang2011sparse} to the application of deep convolutional neural network (CNN) \cite{sun2014deep,sun2014deep2,taigman2014deepface}. While achieving impressive recognition accuracy in controlled environments (some of them even surpass the human performance at certain tasks), most of them also show strong robustness to occlusion and illumination. However, these algorithms largely depend on well-aligned training and testing samples. Research \cite{shan2004curse} has demonstrated that even slight misalignment can globally transform the entire images, greatly reducing the recognition accuracy. Even the CNN that achieves the state-of-the-art performance nowadays needs to align the training and testing faces to the same position, since misaligned query faces can greatly degrade its performance \cite{schroff2015facenet}. Thus current face recognition techniques can benefit from an efficient and well-performing face alignment algorithm.
\par
In this paper, we only consider the face alignment methods based on subspace learning and sparse representation \cite{huang2008simultaneous,yang2012efficient,wagner2012toward}. Although there are other types of face image registration methods that can handle larger face variation in expression and pose, e.g. active appearance models \cite{cootes2001active}, active shape models \cite{cootes1995active} and unsupervised joint alignment \cite{huang2007unsupervised}, their complexity is usually too high for efficient alignment while ours is far more efficient and suitable for real-time situations. Besides, the facial landmarks based methods only focus on accurately detect the facial key points, while ours aligns the face based on the whole training samples and focus on benefiting the subsequent recognition.
\begin{figure}[!t]
\centering
\renewcommand{\captionlabelfont}{\footnotesize}
\setlength{\abovecaptionskip}{4pt}
\setlength{\belowcaptionskip}{-12pt}
\includegraphics[width=3.37in]{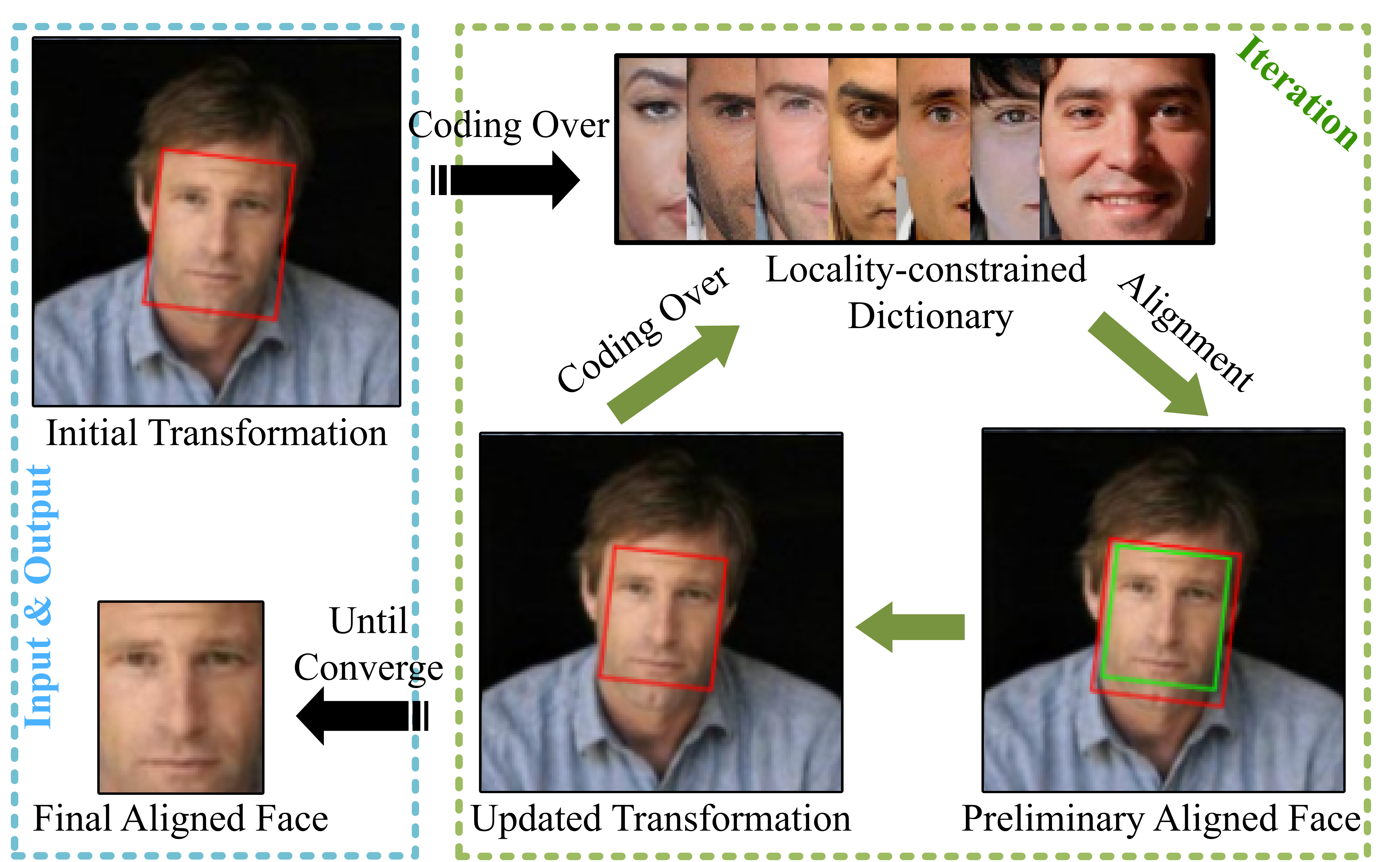}
\caption{\footnotesize . An Illustration of the MRLR. The red bounding box denotes the input estimate and the green one is the output estimate. The whole algorithm works in an iterating style.}\label{fig1}
\end{figure}
 \subsection{Related Work}
 \cite{wright2009robust} reported the sparse representation based classification (SRC), which seeks to represent an aligned testing image by the linear sparse combination of training images. The basic assumption for SRC is that all the training and testing samples need to be well aligned, so SRC performs poorly with misaligned faces. To overcome such shortcoming, \cite{huang2008simultaneous} proposed the transform-invariant sparse representation (TSR). They add deformations in training set, simultaneously recovering the image transformation and representation coefficients. However, TSR aligns testing image to global dictionary and thus easily gets trapped in local minima. To avoid that, robust alignment by sparse representation (RASR) \cite{wagner2012toward} aligns the testing image to training samples of each subject, then warps training set and testing image to a unified transformation for recognition. The exhaustive subject by subject search effectively finds the global optima, but it is extremely time-consuming, especially when the subject number is large. Therefore RASR detrimental to efficiency and scalability. \cite{yang2012efficient} proposed the efficient misalignment-robust representation (MRR) for face recognition. With the carefully controlled training set, they perform the singular value decomposition (SVD) and use principal components to approximate the global dictionary, significantly enhancing its real-time ability. However, SVD operation therein is still time and memory consuming, preventing MRR from being applied in large-scale datasets. Using the principle components of dictionary instead of the original one inevitably reduces alignment accuracy.
 \begin{figure*}[!t]
\centering
\renewcommand{\captionlabelfont}{\footnotesize}
\setlength{\abovecaptionskip}{4pt}
\setlength{\belowcaptionskip}{-12pt}
\includegraphics[width=5.25in]{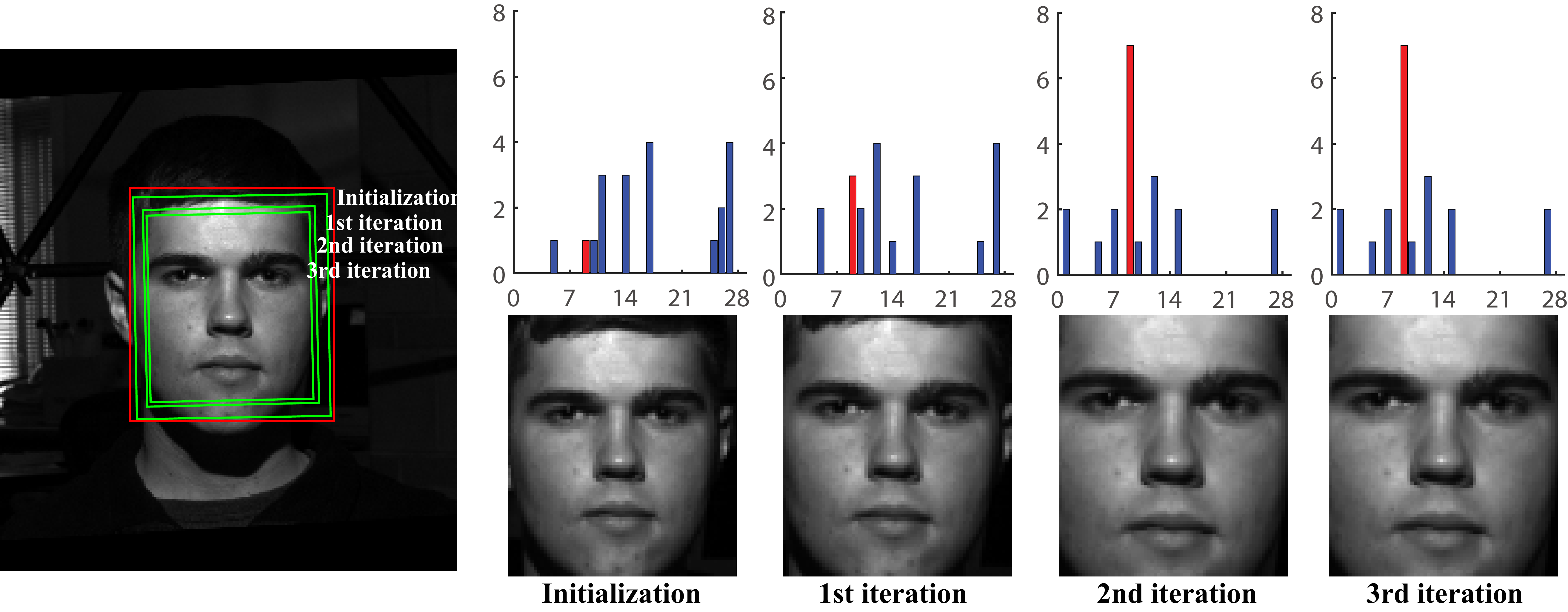}
\caption{\footnotesize . An Illustration of the MRLR iteration procedure. The left image is the input uncropped face. We use the Viola-Jones detector to generate an initialized estimate, and obtain the locality-constrained dictionary (LCD) for it, while the LCD is to compute the new transformation. Then we iteratively update the face transformation and the corresponding LCD until convergence. The bar plot denotes the label distribution of the LCD. It shows that the LCD contains increasingly more training samples from the same class as the testing face after each iteration.}\label{faceiter}
\end{figure*}
\subsection{Motivations and Contributions}
In summary, current prevailing sparse representation based face alignment methods contain several major shortcomings:
\begin{itemize}
\item Time-consuming: \cite{wagner2012toward,zhuang2013single} need to align the face in an exhaustive manner using sub-dictionaries that are constructed by every individual. Suppose the dataset contains more than a thousand individuals, these algorithms will work extremely slow.
\item Easy to introduce background noise: \cite{wagner2012toward,zhuang2013single} align the training set to the testing sample (e.g. \cite{wagner2012toward,zhuang2013single}), which may introduce background noise if the testing sample is largely off-centered and break the low-dimensional linear illumination model \cite{basri2003lambertian}.
\item Unsatisfactory subspace: \cite{huang2008simultaneous,yang2012efficient} use the global dictionary to perform the alignment. The global dictionary contains various uncorrelated face samples and produces a unsatisfactory subspace for alignment.
\item Unable to benefit from outside data.
\end{itemize}
\par
In order to address the above problems, we propose a misalignment-robust locality-constrained representation (MRLR) for robust face recognition. Fig. \ref{fig1} briefly illustrates the MRLR. Inspired by the locality-constrained linear coding \cite{wang2010locality}, the locality is introduced to the dictionary construction for alignment. Specifically, we combine a locality adaptor to the $l_2$ regularized penalties for $\bm{x}$. Because we also use $l_2$ norm to constrain $\bm{e}$, an efficient analytical solution can be derived. While updating the face transformation, we simultaneously update the locality-constrained dictionary, as shown in Fig. \ref{faceiter}. Our contributions are summarized as follows.
\begin{itemize}
\item The proposed locality-constrained representation avoid the exhaustive search in every subject of the training set, greatly reducing the computational time and making the alignment scalable to large datasets. To the best of our knowledge, this is the first time that locality has been introduced to improve the performance of face alignment.
\item MRLR uses the locality adaptor and the $l_2$-norm to penalize the the representation term and the error term. We derive an analytical solution for the optimization. Moreover, we can accelerate the analytical solution by making use of the block structure of the deformable dictionary. Thus the inverse of a large-size matrix can be further avoided, making our model even more scalable and efficient.
\item MRLR simultaneously optimize the transformation and update the corresponding locality-constrained dictionary, which largely avoids the unsatisfactory local minima.
\item MRLR can take advantage of outside data to better construct the locality-constrained dictionary. Outside data can be effectively used to benefit the alignment performance.
\end{itemize}
\begin{figure*}[!t]
\centering
\renewcommand{\captionlabelfont}{\footnotesize}
\setlength{\abovecaptionskip}{2pt}
\setlength{\belowcaptionskip}{-10pt}
\includegraphics[width=6.2in]{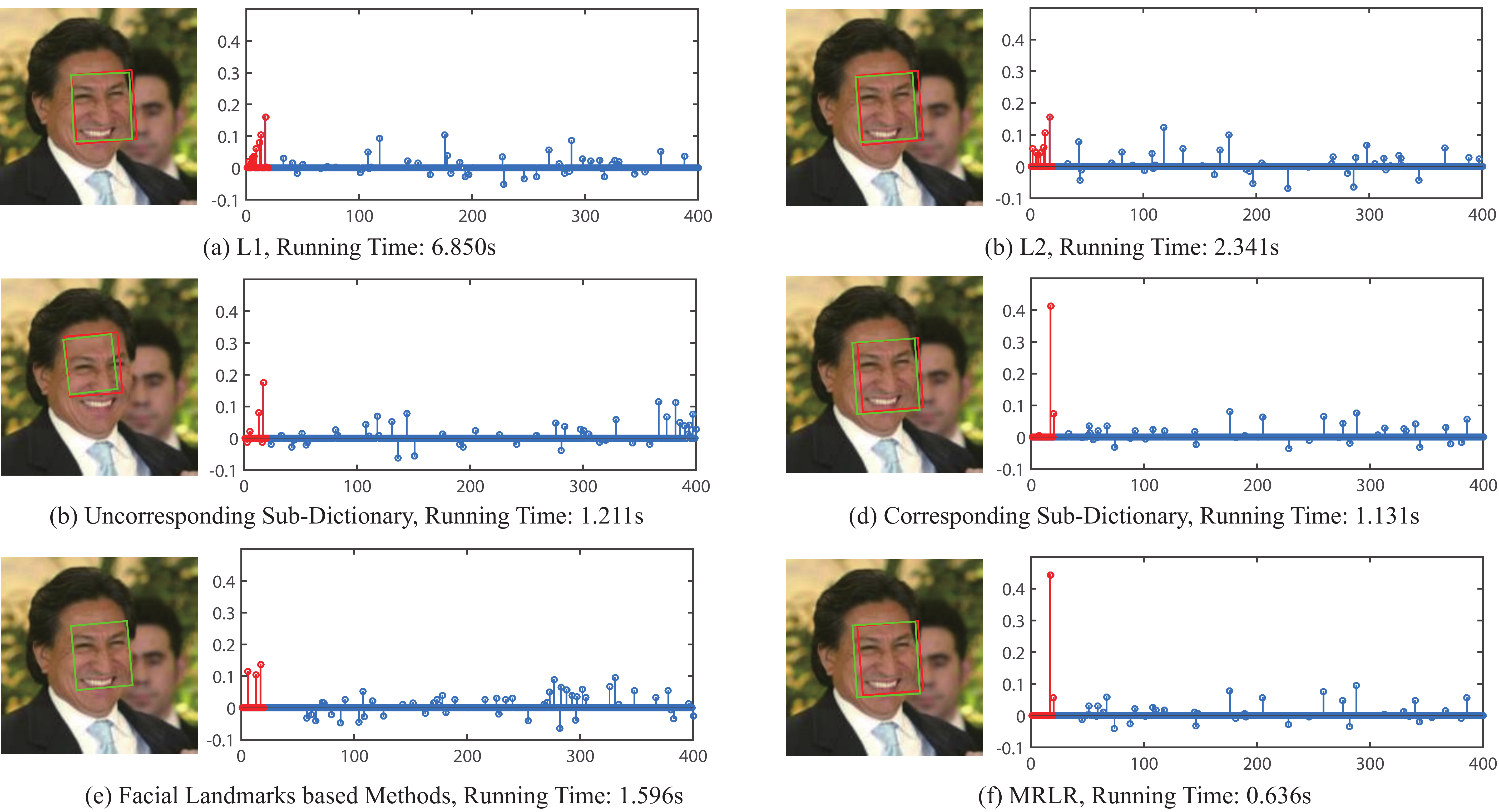}
\caption{\footnotesize . A face alignment example using different constraints, key point based method and MRLR. The red box is the initial estimate and the green box denotes the final alignment. After performing face alignment, we use the SRC to perform the face recognition. The stem diagram on the right shows the corresponding sparse representation coefficients of the aligned face after performing SRC. We can see only (d) and (f) show discriminative representation results. Note that, the corresponding sub-dictionary in (d) means we only use the sub-dictionary that is constituted by faces whose label is the same as the testing face. The uncorresponding sub-dictionary in (c) is constituted by faces that do not belong to the testing face.}\label{fig2}
\end{figure*}
\section{The Proposed Method}
\subsection{The MRLR Model}
We arrange the given $n_i$ training samples from the $i$th class as columns of a dictionary $\bm{D}_i=[\bm{d}_{i,1},\bm{d}_{i,2},\cdots,\bm{d}_{i,n_i}]\in \mathbb{R}^{m\times n_i}$ where $\bm{d}_{i,j}\in\mathbb{R}^m$ denotes the $j$th vectorized training sample of the $i$th class. Combining all the dictionary from each subjects, we can obtain a global dictionary $\bm{D}=[\bm{D}_1,\bm{D}_2,\cdots,\bm{D}_k]\in\mathbb{R}^{m\times n}$ where $n=\sum_{i=1}^k n_i$ and $k$ is the number of subjects. Suppose that the query face $\bm{y}$ belongs to the $i$th subject, ideally it can be approximately represented by $\bm{D}_i\bm{x}$ in which $\bm{x}$ is the representation coefficients. However, due to the misalignment problem, such linear subspace representation may be invalid. Therefore we introduce a transformation that models the warping to the original face. Instead of observing the $\bm{y}$, we observe the warped face $\bm{y}_w=\bm{y}\circ\tau^{-1}$ where $\circ$ denotes a nonlinear operator and $\tau$ belongs to a finite-dimensional group of transformations acting on the image domain (e.g. similarity transformation). The linear subspace representation $\bm{x}$ of the warped face can not reveal the true identity. Naively applying recognition algorithm is inappropriate. On the other hand, the potential subspace corresponding to $\bm{y}$ is also unknown, so it is difficult to align it. Fortunately, by leveraging the high similarity of face, we can construct a suboptimal local dictionary for alignment, and update the local dictionary according to the latest transformation. After several iterations, it eventually converges to the accurate transformation. After the true deformation $\tau^{-1}$ is found, then we can apply its inverse $\tau$ to the testing face and obtain the aligned face $\bm{y}_w\circ\tau$.
\par
The global dictionary with $l_1/l_2$ constraint usually recovers the unsatisfactory transformation (see Fig. \ref{fig2}), because it is prone to local minima under the interference of atoms from the other subjects. Inspired by \cite{wang2010locality}, we introduce the locality constraints to the dictionary. The reason lies in two folds. First, locality-constrained dictionary only uses the most similar atoms to the query, effectively avoiding unsatisfactory local minima caused by dissimilar atoms, as shown in Fig. \ref{fig2}. Second, using locality-constrained dictionary  requires no exhaustive search in every subject and leads to highly efficient solving algorithm. The model of MRLR is formulated as
\begin{equation}\label{MRLRm}
\small
\min_{\bm{x},\bm{e}}\ \|\bm{c} \odot \bm{x}\|_{2}^{2}+\|\bm{e}\|_{2}^{2}\quad\ \textnormal{s.t.}\ \ \bm{y}_{w}\circ \tau=\bm{D}\bm{x}+\bm{e}
\end{equation}
where $\odot$ denotes the element-wise multiplication between two vectors, and $\bm{c}\in\mathbb{R}^{n}$ is the locality adaptor that attaches different penalties to the coefficients $\bm{x}$. The locality adaptor activates the most correlated atoms for the testing sample, while suppressing the uncorrelated ones. Unfortunately, the model in Eq. \eqref{MRLRm} is difficult to solve due to the non-linearity. A small deformation in the transform $\tau$ can be linearized as $\bm{y}_{w}\circ (\tau + \Delta \tau)=\bm{y}_{w}\circ \tau+ J\Delta \tau$ where $J=\frac{\partial }{\partial r} \bm{y}_{w}\circ \tau$ is the Jacobian of $\bm{y}\circ \tau$ with respect to $\tau$ and $\Delta \tau$ is the step in $\tau$.  If an initial $\tau$ is given, we can repeatedly search for an optimal $\Delta \tau$ to update $\tau$ and $J$. A final transformation $\tau$ can be obtained to align the warped image.
\par
The efficiency of the MRLR model lies in two folds. First, we enforce the $l_2$ norm constraints on both $\bm{c}\odot \bm{x}$ and $\bm{e}$, and derive an analytical solution for MRLR, which is much faster than solving $l_1$ norm minimization. In fact, the performance of the $l_2$ constraints are similar to the $l_1$ constraints in the case without occlusion \cite{zhang2011sparse}. Second, we take advantage of the block structure of matrices to design a highly efficient algorithm, which obtains exactly the same solution in shorter time. The MRLR algorithm is summarized as follows.
{
\begin{algorithm}[h]
\small
    \caption{The MRLR algorithm for Face Alignment}
    \label{alg1}
    \begin{algorithmic}[1]
    \REQUIRE~The dictionary of training samples $\bm{D}$, the warped testing image $\bm{y}_w$, the initial transformation $\tau$ (it can be obtained by any off-the-shelf face detector, e.g. Viola-Jones detector), a constant $\sigma$.
    \ENSURE~The aligned face $\bm{y}$\\
    \WHILE {not converge or reach maximal iteration}
    \STATE Compute the locality adaptor: $\bm{c} \leftarrow \exp(\frac{\bm{D}^T\bm{y}}{\sigma})$, for all $i$, $\bm{c}_i \leftarrow max(\bm{c})-\bm{c}_i$.
    \STATE $j\leftarrow 1$.
    \WHILE {not converge or reach maximal iteration}
    \STATE $\hat{\bm{y}}_{w}(\tau_{j-1})\leftarrow\frac {\bm{y}_{w}\circ \tau_{j-1}}{\|\bm{y}_{w}\circ \tau_{j-1}\|_{2}}$, $J\leftarrow\frac {\partial}{\partial\tau_{j-1}}\hat{\bm{y}}_{w}(\tau_{j-1})|_{\tau_{j-1}}$.
    \STATE $\begin{aligned}
& \Delta \tau=\arg\min \limits_{\Delta \tau,\bm{x},\bm{e}}\ \|\bm{c} \odot\boldsymbol{x}\|_{2}^{2}+\|\bm{e}\|_{2}^{2}\\
& \textnormal{s.t.}\quad \hat{\bm{y}}_{w}(\tau_{j})+J\Delta\tau=\bm{D}\bm{x}+\bm{e}
\end{aligned}$.
    \STATE $\tau_j\leftarrow\tau_{j-1}+\Delta \tau$.
    \STATE $j\leftarrow j+1$.
    \ENDWHILE
    \STATE $\tau\leftarrow \tau_j$, $\tau_0\leftarrow \tau_j$.
    \ENDWHILE
    \STATE Output the final aligned face $\bm{y}=\bm{y}_w\circ\bm{e}$.
    \end{algorithmic}
 \end{algorithm}
 }
\subsection{Efficient Solving Algorithm}
This section presents a highly efficient solution for the MRLR algorithm. By analyzing Algorithm \ref{alg1}, we find the optimization in step 6 dominates the overall computational time. Although it has an analytical solution, it contains the inversion operation of a large-size matrix. We aim to take advantage of the block structure of the matrix to decompose the inversion. We first reformulate the optimization in Step 6 as
\begin{equation}\label{reMRLR}
\small
\begin{aligned}
&\Delta \bm{\tau} = \min \limits_{\Delta \tau}\|\bm{C} \bm{x}\|_{2}^{2}+\|\bm{e}\|_{2}^{2}\\
&\ s.t.\ \ \hat{\bm{y}}_{w} + J\Delta \tau  =\bm{D}\bm{x}+\bm{e}
\end{aligned}
\end{equation}
where $\bm{C}$ is a diagonal matrix with the diagonal elements being the locality adaptor vector $\bm{c}$. We can further substitute $
\bm{e}=\hat{\bm{y}}_{w}-\left[ \bm{D},-J\right]
\left[
  \begin{array}{c}
    \bm{x} \\
    \Delta \tau \\
  \end{array}
\right]
$ into Eq. \eqref{reMRLR}, we have
\begin{equation}\label{MRLR2}
\small
\begin{aligned}
& \Delta \tau=\arg\min \limits_{\bm{x},\varDelta \tau}\ \|\bm{C}\bm{x}\|_{2}^{2}+
\left\|\hat{\bm{y}}_{w}-[\bm{D},-J]
\left[
  \begin{array}{c}
    \bm{x} \\
    \Delta\tau \\
  \end{array}
\right]
\right\|_{2}^{2}    \\
& \quad \  =\arg\min \limits_{\bm{x},\Delta \tau}
\left\|
\left[
  \begin{array}{c}
    \hat{\bm{y}}_{w} \\
    \bm{0} \\
  \end{array}
\right]
-
\left[
  \begin{array}{cc}
    \bm{D} & -J \\
    \bm{C} & \bm{0} \\
  \end{array}
\right]
\left[
  \begin{array}{c}
    \bm{x} \\
    \Delta\tau \\
  \end{array}
\right]
\right\|_{2}^{2}    \\
& \quad \  =\arg\min \limits_{\bm{z}}
\left\|
\bm{u}-R\bm{z}
\right\|_{2}^{2}
\end{aligned}
\end{equation}
where $\bm{u}$, $\bm{R}$ and $\bm{z}$ denote $
\left[
  \begin{array}{c}
    \hat{\bm{y}}_{w} \\
    \bm{0} \\
  \end{array}
\right]
$, $\left[
  \begin{array}{cc}
    \bm{D} & -J \\
    \bm{C} & \bm{0} \\
  \end{array}
\right] $ and $\left[
  \begin{array}{c}
    \bm{x} \\
    \Delta\tau \\
  \end{array}
\right]$ respectively. It becomes a least square problem whose analytical solution is $\bm{z}=(\bm{R}^{T}\bm{R})^{-1}\bm{R}^{T}\bm{u} $. As one can see, the computational complexity is still high due to the large size of $\bm{R}$. Actually, the efficiency and the scalability can be greatly boosted if we make good use of the block structure of the matrix $\bm{R}$.
\par
Using the block matrix inversion, we can rewrite the analytical solution as
\begin{equation}\label{solutionz}
\small
\begin{aligned}
& \boldsymbol{z}\ =(\bm{R}^{T}\bm{R})^{-1}\bm{R}^{T}\bm{u} \\
& \quad  =\left(\left[\!\!
  \begin{array}{cc}
    \bm{D}^{T}\! &\!\! \bm{C}^T \\
    -J^{T}\! &\!\! \bm{0} \\
  \end{array}
\!\!\right]
\left[
  \begin{array}{cc}
    \bm{D}\! &\!\! -J \\
    \bm{C}\! &\!\! \bm{0} \\
  \end{array}
\right]\right)^{-1}\!
\left[
  \begin{array}{cc}
    \bm{D}^{T}\! &\!\! \bm{C} \\
    -J^{T}\! &\!\! \bm{0} \\
  \end{array}
\right]
\left[
  \begin{array}{c}
    \hat{\bm{y}}_{w}\!\! \\
    \bm{0} \\
  \end{array}
\right] \\
& \quad  =
\left[\!\!
  \begin{array}{cc}
    \bm{D}^{T}\bm{D}+\bm{C}^{T}\bm{C} & -\bm{D}^{T}J \\
    -J^{T}\bm{D} & J^{T}J \\
  \end{array}
\!\!\right]
^{-1}\!
\left[\!\!
  \begin{array}{c}
    \bm{D}^{T} \\
    -J^{T} \\
  \end{array}
\!\!\right]\hat{\boldsymbol{y}}_{w} \\
& \quad  =
\left[\!\!
  \begin{array}{cc}
    \bm{Z}_{1}^{-1} \!\!\!& \!\!\!\!
    \begin{aligned}
    &\!\!\!\!(\bm{D}^{T}\bm{D}+\bm{C}^{T}\bm{C})^{-1}\\
    &\!\!\!\!\ \ \ \times(\bm{D}^{T}J)\bm{Z}_{2}^{-1}
    \end{aligned}
    \\
    \begin{aligned}
    &\ \bm{Z}_{2}^{-1}(J^{T}\bm{D})\times\!\!\!\!\\
    &(\bm{D}^{T}\bm{D}+\bm{C}^{T}\bm{C})^{-1}\!\!\!\!
    \end{aligned}\ \!\!\!\!
    & \!\!\!\!\bm{Z}_{2}^{-1}\\
  \end{array}
\!\!\right]
\left[
  \begin{array}{c}
    \!\!\bm{D}^{T} \!\!\!\!\\
    \!\!-J^{T} \!\!\!\!\\
  \end{array}
\right]\hat{\bm{y}}_{w} \\
\end{aligned}
\end{equation}
We denote $\bm{D}^{T}\bm{D}+\bm{C}^{T}\bm{C}$, $\bm{D}^TJ$ and $J^TJ$ as $\bm{T}_1$, $\bm{T}_2$ and $\bm{T}_3$ respectively. In particular, $\bm{T}_1$ and $\bm{T}_1^{-1}$ can be pre-calculated before the inner iteration from Step 4 to Step 9. The other variables $\bm{Z}_{1}$ and $\bm{Z}_{2}$ can be represented as $\bm{Z}_{1}^{-1}=(\bm{T}_{1}-\bm{T}_{2}\bm{T}_{3}^{-1}\bm{T}_{2}^{T})^{-1} $ and $\bm{Z}_{2}^{-1}=(\bm{T}_{3}-\bm{T}_{2}^{T}\bm{T}_{1}^{-1}\bm{T}_{2})^{-1}$. Eq. \eqref{solutionz} can be represented as
\begin{equation}
\small
\bm{z}
=
\left[
  \begin{array}{c}
    \bm{x} \\
    \Delta \tau \\
  \end{array}
\right]
=
\left[
  \begin{array}{c}
    \bm{Z}_{1}^{-1}(\bm{D}^{T}\hat{\bm{y}}_{w})-\bm{T}_{1}^{-1}\bm{T}_{2}\bm{Z}_{2}^{-1}(J^{T}\hat{\bm{y}}_{w}) \\
    \bm{Z}_{2}^{-1}\bm{T}_{2}^{T}\bm{T}_{1}^{-1}(\bm{D}^{T}\hat{\bm{y}}_{w})-\bm{Z}_{2}^{-1}(J^{T}\hat{\bm{y}}_{w}) \\
  \end{array}
\right]
\end{equation}
\par
Note that the purpose of the face alignment is to search a deformation step $\Delta \tau$, so computing $\bm{x}$ is unnecessary. Without computing $\bm{x}$, we can save greatly reduce the computation. Moreover, as mentioned in \cite{wang2010locality}, since $\bm{c}$ usually imposes weak constraint on only a few atoms, suppressing most of the atoms. We can simply keep the smallest $s,(s\ll n)$ entries in $\bm{c}$ and force other entries to be positive infinity. This strategy further accelerates the coding, as we present in complexity analysis and experiments (This strategy is termed as MRLR2, while the former proposed one is termed as MRLR1). Detailed complexity analysis refers to the supplementary material 
\begin{figure*}[t]
\centering
\renewcommand{\captionlabelfont}{\footnotesize}
\setlength{\abovecaptionskip}{2pt}
\setlength{\belowcaptionskip}{-10pt}
\includegraphics[width=6.95in]{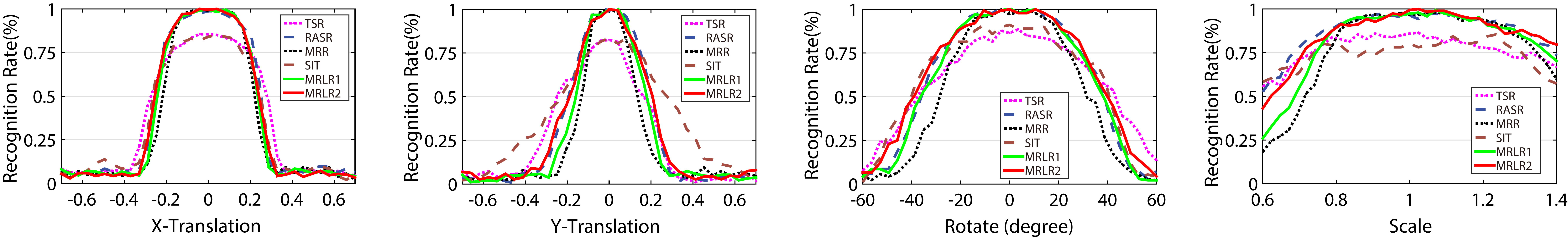}
\caption{\footnotesize . The region of attraction. (The amount of translation is given as a fraction of the distance between eyes) (a) Translation in the Y-direction only. (b) Translation in the X-direction only. (c) In-plane rotation only. (d) Scale variation only.}\label{exp1}
\end{figure*}
\section{Experiments}
We conduct experiments on the face database (Extended Yale B \cite{georghiades2001few} and CAS-PEAL \cite{gao2008cas}) with controlled laboratory conditions to comprehensively evaluate MRLR in terms of region of attraction, recognition rate, running time and scalability. Then practical face recognition performance are evaluated by the Labeled Faces in the Wild (LFW) dataset \cite{huang2007labeled}. The experimental results show that MRLR achieves competitive performance with much less running time and scales better in large datasets. Moreover, MRLR is able to make use of outside data to improve alignment, benefiting the subsequent recognition in the scenario where only one sample each person is available.
\subsection{Implementation details}
In MRLR2 and MRR, the length (the number of atoms) of dictionary for alignment is fixed to 20 for fair comparison. We basically follow the same setting in \cite{wagner2012toward}, 10 classes after first stage are remained in MRR and RASR, and one project matrix of 500 rows is used in TSR. The illumination dictionary in \cite{zhuang2013single} follows its original setup, and the amount of illumination atoms is 30 in all experiments. The maximum iteration of outer and inner loop for MRLR in these methods are consistently set to 3 and 30.. The $l_1$-minimization algorithm uses the Augmented Lagrange Multiplier \cite{yang2010fast}.
\subsection{The region of attraction}
The region of attraction evaluates the robustness against 2D deformations. We compare MRLR with TSR \cite{huang2008simultaneous}, RASR \cite{wagner2012toward}, MRR \cite{yang2012efficient} and SIT \cite{zhuang2013single} on Extended Yale B database, which includes 2414 images of 38 subjects. We use the uncropped images of 28 subjects in experiments. 32 training images per category are randomly selected, and the rest are used for testing. All the training images are resized to $80\times 70$. We get access to the ground truth of eyes and add perturbation to them. Then we calculate the corresponding recognition accuracy under various initial transformations. One can see that MRLR performs well and stably within a certain range of misalignment, e.g. 20 percent translation in x direction (14-16 pixels), 20 degree rotation or 30 percent scale variation. It significantly enhances the robustness of practical recognition, because the average misalignment of a face detector safely falls within 10 percent translation and 8 percent scale variation. TSR performs relatively poor even in small perturbations. It is mainly because aligning testing image to the entire training set is more prone to local minima, resulting in inaccurate alignment. Using single sample per class, SIT achieves similar performance with TSR due to the limited representation ability. Compared to MRLR1, MRLR2 and RASR, MRR performs slightly worse on robustness to deformation. MRLR1 and MRLR2 perform almost the same as RASR, demonstrating that locality-constrained representation effectively avoids local minima.
\subsection{Face recognition in controlled environments}
\begin{table}[!b]
\centering
\renewcommand{\captionlabelfont}{\footnotesize}
\setlength{\abovecaptionskip}{4pt}
\setlength{\belowcaptionskip}{-8pt}
\qihao
\caption{\footnotesize . The recognition accuracy and running time on Extended Yale B and CAS-PEAL datasets.}\label{tab1}
\begin{tabular}{|c|c|c|c|c|}\hline
\bb{Method} & \multicolumn{2}{c|}{Extended Yale B} & \multicolumn{2}{c|}{CAS-PEAL}\\
\cline{2-5}
& Recognition & Running  & Recognition  & Running\\
& Rate &  Time &Rate &  Time\\\hline\hline
TSR	& 81.61 & 7.396 & 86.96 & 4.2695\\
RASR & 92.42 & 9.7587 & 89.92 & 5.4466\\
MRR	& 90.95 & 0.7773 & 90.00 & 0.5684\\
SIT	& 84.53 & 9.9823 & 86.76 & 6.0329\\\hline
\textbf{MRLR-SS} & 77.12 & \textbf{0.1566} & 81.31 & \textbf{0.1384}\\
\textbf{MRLR1} & 92.31 & 0.6207 & 89.76 & 0.3307\\
\textbf{MRLR2} & \textbf{92.53} & 0.1783 & \textbf{90.43} & 0.1462\\\hline
\end{tabular}
\end{table}
We conduct the recognition experiments on both Extended Yale B and CAS-PEAL datasets. For Extended Yale B, we adopt the same settings in the previous section. For CAS-PEAL, 20 subjects were chosen, each of them including more than 32 images. We randomly selected 20 images per subject and resized them to $80 \times 70$ for training, then test on the remaining 12 images. Because SIT \cite{zhuang2013single} trains on single sample per category, we also reduce the training set in MRLR to single sample per category for fair comparison (termed as MRLR-SS). The initial $\tau_0$ are automatically given by Viola-Jones detector \cite{viola2001rapid}. Table \ref{tab1} gives the recognition rates and average running time.
\par
As we discuss above, unsatisfactory local minima in global dictionary leads to poor performance (81.61\%) in TSR. With less amount of subject (from 32 to 20), local minima is alleviated and TSR is able to perform better. RASR performs very well in both Extended Yale B dataset (92.42\%) and CAS-PEAL dataset (89.92\%). However, such subject by subject search is time-consuming, it averagely costs 9.76 and 5.45 seconds on each testing image when the amount of subjects are 28 and 20, respectively. On the other hand, the recognition rate of MRLR2 is 92.68\% and 90.43\%, slightly better than RASR. Most importantly, it takes only 0.18 and 0.15 seconds to deal with a testing image, roughly 4, 55 and 41 times faster than MRR, RASR and TSR respectively. With single sample each subject, SIT achieves 84.53\% and 86.76\% recognition rate in two datasets respectively, better than the single sample version of MRLR. Because the dictionary for alignment consists of illumination dictionary (outside samples) and single training sample per class, it shares the same scale with RASR, resulting in similar running time.
\subsection{Scalability}
We vary the number of subject from 10 to 100 and resize the images from $40\times 35$ to $160\times 140$, to evaluate the scalability of our algorithm. Table \ref{tab2} and Table \ref{tab3} show the experimental results. One can observe that TSR, RASR and SIT cost too much time, far from being applicable in real-time systems. The running time of TSR remains relatively stable as the dimension increases, but rises linearly with more subjects. MRR maintains excellent real-time capability with the growth of the subject number. However, its running time rises dramatically when the resolution of image increasing. Unlike the abovementioned approaches, MRLR1 and MRLR2 are not very sensitive to the dimension or number of subjects, preserving competitive performance. MRLR2 costs the least running time and the lowest increasing rate as we enlarge the dimension or number of subjects, showing the best scalability among state-of-the-art approaches.
\begin{table}[!h]
\renewcommand{\captionlabelfont}{\footnotesize}
\setlength{\abovecaptionskip}{4pt}
\setlength{\belowcaptionskip}{-4pt}
\centering
\qihao
\caption{\footnotesize . Running time (s) under different dimensions (image size).}\label{tab2}
\begin{tabular}{|c|c|c|c|c|c|}
\hline
Method & $\!40\!\times\! 35\!$ & $\!64\!\times\! 56\!$ & $\!80\!\times\! 70\!$ & $\!120\!\times\! 105\!$ & $\!160\!\times\! 140\!$\\\hline\hline
TSR & 3.645 & 3.861 & 4.270 & 4.672 & 5.468\\
RASR & 3.499 & 4.452 & 6.110 & 10.324 & 17.111\\
MRR & 0.133 & 0.342 & 0.593 & 2.259 & 5.997\\
SIT & 3.564 & 4.637 & 6.565 & 11.035 & 19.215\\\hline
\textbf{MRLR1} & 0.085 & 0.195 & 0.331 & 0.569 & 0.940\\
\textbf{MRLR2} & \textbf{0.066} & \textbf{0.118} & \textbf{0.146} & \textbf{0.303} & \textbf{0.505}\\\hline
\end{tabular}
\end{table}
\begin{table}[!h]
\renewcommand{\captionlabelfont}{\footnotesize}
\setlength{\abovecaptionskip}{4pt}
\setlength{\belowcaptionskip}{-7pt}
\centering
\qihao
\caption{\footnotesize . Running time (s) under different amount of classes.}\label{tab3}
\begin{tabular}{|c|c|c|c|c|c|}
\hline
Method & 10 & 20 & 40 & 70 & 100\\\hline\hline
TSR & 2.1533 & 3.2825 & 5.5280 & 8.4034 & 11.5327\\
RASR & 2.7377 & 4.6596 & 8.8647 & 15.4644 & 22.1281\\
MRR & 0.5776 & 0.5928 & 0.6082 & 0.6394 & 0.6994\\
SIT & 2.86 & 5.1996 & 9.9817 & 17.6875 & 27.1734\\\hline
\textbf{MRLR1} & 0.1977 & 0.2819 & 0.513 & 0.8552 & 1.4096\\
\textbf{MRLR2} & \textbf{0.1318} & \textbf{0.1373} & \textbf{0.1559} & \textbf{0.197} & \textbf{0.2616}\\\hline
\end{tabular}
\end{table}
\subsection{Face recognition and verification in the wild}
In this section, we test MRLR in practical scenario. LFW dataset contains 13,233 images of 5,749 people, while 4,069 people have only one image. This dataset is very challenging since it is collected in the uncontrolled wild scenario, including blur, various illumination, crossing age, occlusion or misalignment. We present two experiments, face recognition and face verification on LFW database to evaluate the performance of MRLR. In recognition testing, we choose 20 persons with more than 20 images, forming a subset with 1534 samples. We randomly select 20 samples each subject as training set, and test on the rest. The experimental results are shown in Table 3. It is worth noticing that there are many testing images including 3D deformation. Although aligning a 3D warped image to frontal face is beyond the scope of our approach, we do not manually exclude these images, for the purpose of evaluating the performance of our method in practical scenario. It is clear that MRLR performs best among these misalignment-robust recognition algorithms, outperforming RASR for 1.55\% in recognition rate. Furthermore, the single sample version of MRLR beats SIT with a significant margin. It is mainly because the illumination dictionary is not informative enough to represent such sophisticated intra-class variation in each subject.
\par
To address the problem of insufficient training images when aligning, we propose to use outside data to improve alignment. Making fully use of the similarity of face, the outside data that belong to neither the training subjects nor the testing subjects, also enhances the accuracy of alignment. With the outside data, MRLR performs better even in the scenario where there is only one sample per subject.
\begin{table}[!h]
\renewcommand{\captionlabelfont}{\footnotesize}
\setlength{\abovecaptionskip}{4pt}
\setlength{\belowcaptionskip}{-7pt}
\centering
\qihao
\caption{\footnotesize . Recognition accuracy (\%) on LFW dataset.}\label{tab4}
\begin{tabular}{|c|c|c|c|}
\hline
Method & Accuracy & Method & Accuracy\\\hline\hline
TSR & 73.63 & \textbf{MRLR-SS} & 72.47\\
RASR & 81.43 & \textbf{MRLR-SS} with outside data & 80.42\\
MRR & 78.84 & \textbf{MRLR1} & \textbf{82.98}\\
SIT & 55.91 & \textbf{MRLR2} & 81.75\\\hline
\end{tabular}
\end{table}
\par
Face verification represents another task. Given two face images, the goal is to decide whether the two people pictured belong to the same individual. Many breakthroughs have been achieve by Convolutional Neural Network (CNN) \cite{krizhevsky2012imagenet,simonyan2014very}. However, the training data also need to be loosely or accurately aligned, so that the verification accuracy can be further boosted. In \cite{sun2014deep,sun2014deep2,sun2014deeply}, similarity transformation is used to align training and testing images according the facial landmarks. \cite{taigman2014deepface,schroff2015facenet} also state that accurate 3D aligning do help the subsequent face verification. In this experiment, we train a simple neural network consisting of 7 convolution layers and 2 fully connected layers, jointly supervised by softmax loss and contrastive loss. The specific deep model description is given in the supplementary material. The deep model is trained on roughly 600 thousand outside samples (These samples and LFW do not share the same individuals.) and test on LFW, following the standard unrestricted protocol. The feature of each image are taken from the output of the first fully connected layer, and their Euclidean distance are calculated for binary classification. The pairs whose distance exceeds the threshold are regarded as negative. We compare landmarks based (IntraFace \cite{asthana2014incremental}) method and MRLR by aligning the testing image, and carry out ten-fold cross validation testing on 6000 pairs. The results are reported in Table \ref{tab5}.
\begin{table}[!h]
\renewcommand{\captionlabelfont}{\footnotesize}
\setlength{\abovecaptionskip}{4pt}
\setlength{\belowcaptionskip}{-7pt}
\centering
\qihao
\caption{\footnotesize . Verification accuracy (\%) on LFW dataset.}\label{tab5}
\begin{tabular}{|c|c|c|c|}
\hline
Method & No. of points & Distance & Accuracy (\%)\\\hline\hline
Intraface & 5 & L2 & 98.05$\pm$0.64\\
Intraface & 5 & PCA+L2 & 98.00$\pm$0.68\\
Intraface & 12 & L2 & 98.09$\pm$0.60\\
Intraface & 12 & PCA+L2 & 98.15$\pm$0.49\\
Intraface & 49 & L2 & 98.22$\pm$0.61\\
Intraface & 49 & PCA+L2 & 98.32$\pm$0.47\\\hline
\textbf{MRLR2} & N/A(Image Set) & L2 & \textbf{98.68}$\pm$\textbf{0.51}\\
\textbf{MRLR2} & N/A(Image Set) & PCA+L2 & \textbf{98.77}$\pm$\textbf{0.45}\\\hline
\end{tabular}
\end{table}
\section{Concluding Remarks}
In this paper, we propose an efficient misalignment-robust locality representation algorithm, MRLR, for face alignment. The locality constraint therein avoids the interference of the uncorrelated atoms and the exhaustive search in every subject, greatly reducing running time while still preserving accurate alignment. Moreover, motivated by the block structure of dictionary, we propose an efficient solving algorithm to speed up the alignment. Besides, MRLR is easily extended to one-shot face alignment and can benefit from outside data. Computational complexity analysis and extensive experiments show that MRLR considerably reduce the running time with even better performance.
{
\bibliographystyle{aaai}
\bibliography{MRLR}
}
\end{document}